%% file: main.tex
\documentclass[runningheads]{llncs}

\usepackage{eccv}

\usepackage{eccvabbrv}

\usepackage{graphicx}
\usepackage{booktabs}

\usepackage[accsupp]{axessibility}  

\usepackage{hyperref}

\usepackage{orcidlink}

\usepackage{multirow}
\usepackage{tabularx}
\usepackage{array}
\usepackage[table]{xcolor}
\usepackage{enumitem}
\usepackage{footmisc}

\begin{document}

\title{Multimodal Emotion Recognition in Conversations via Class-Wise Adaptive Modality Fusion and Affective Geometry} 

\titlerunning{Multimodal ERC via Class-Wise Adaptive Modality Fusion}

\author{Oriol Marín\inst{1}\orcidlink{0009-0009-6645-3401} \and
Roger Marí\inst{1}\orcidlink{0009-0004-7236-8004} \and
Gloria Haro\inst{2,3}\orcidlink{0000-0002-8194-8092} \and
Rafael Redondo\inst{1}\orcidlink{0000-0001-9737-8939}}

\authorrunning{O.~Marín et al.}


\institute{Eurecat, Centre Tecnològic de Catalunya, Barcelona, Spain \\
\email{\{oriol.marin,roger.mari,rafael.redondo\}@eurecat.org} \and Universitat Pompeu Fabra, Barcelona, Spain \and
Serra Húnter Fellow Programme, Universitat Pompeu Fabra, Barcelona, Spain \\
\email{gloria.haro@upf.edu}}

\maketitle

\input{sections/0_abstract}
\input{sections/1_introduction}

\input{sections/2_relatedwork}
\input{sections/3_method}
\input{sections/4_experiments}
\input{sections/5_conclusion}


\section*{Acknowledgements}
This work was financially supported by the Catalan Government through the funding grant ACCIÓ-Eurecat (Project TRAÇA: “MentalTwin” 2026-2027).

%
%
\bibliographystyle{splncs04}
\bibliography{main}
\end{document}

%% file: sections/0_abstract.tex
\begin{abstract}

Emotion Recognition in Conversations (ERC) requires integrating heterogeneous textual, audio, and visual cues while accounting for conversational context and emotional dynamics. We extend the Self-Distillation Transformer architecture for ERC with appearance+geometry visual representations, class-wise adaptive modality fusion, and a valence-arousal prior for affective transitions. On the MELD and IEMOCAP datasets, geometry-enhanced visual representations improve weighted F1 by 0.27 and 4.36 points over appearance-only features, respectively, while class-wise adaptive fusion provides further gains of 0.17 and 0.25 points over the original softmax gate. The valence-arousal prior yields targeted improvements of 0.30 and 0.74 accuracy points on emotionally shifted utterances while preserving performance on stable turns. These results indicate that structured facial cues, emotion-dependent modality weighting, and affective geometry provide complementary benefits for multimodal ERC.

  \keywords{Multimodal emotion recognition in conversations \and Multimodal fusion \and Affective geometry \and Affective computing}
\end{abstract}

%% file: sections/1_introduction.tex
\section{Introduction}
\label{sec:introduction}

Emotion Recognition in Conversations (ERC) aims to identify the emotion expressed in each conversational turn, or utterance, by jointly modeling linguistic content, vocal cues, facial behavior, conversational context, and speaker interactions~\cite{poria2019emotion}. Unlike isolated emotion recognition, ERC must account for how affect evolves throughout a dialogue and how the relevance of each modality changes across speakers, emotions, and recording conditions.

\begin{figure}[t]
  \centering
  \includegraphics[width=0.95\columnwidth]{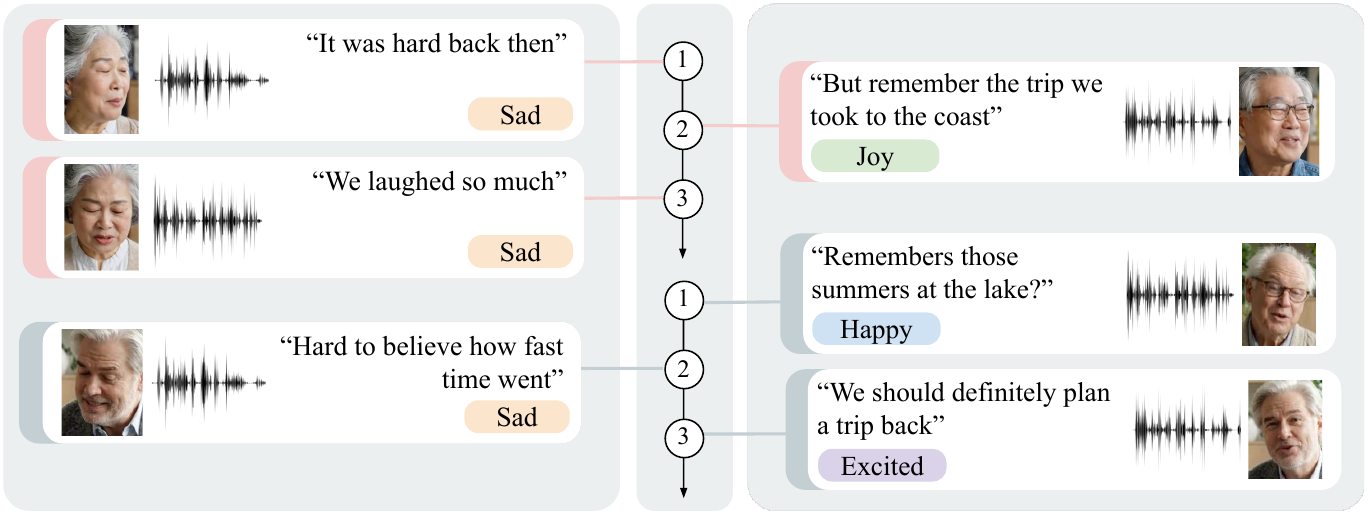}
  \caption{
  Examples of emotion-shift dialogues in multimodal ERC. Top: \textit{sad\,$\to$\,joy\,$\to$\,sad}; bottom: \textit{happy\,$\to$\,sad\,$\to$\,excited}. Each turn shows the speaker face, audio waveform, transcribed utterance, and emotion label, illustrating how affect evolves across turns.
  }
  \label{fig:erc_example}
\end{figure}

Multimodal ERC remains challenging for three main reasons. First, text, audio, and visual signals are heterogeneous and are not equally informative for every utterance. In practice, multimodal models may become dominated by textual representations, while audio and visual cues receive comparatively less influence in standard fusion mechanisms~\cite{peng2022balanced}. Second, the informativeness of each modality varies across utterances and
emotion categories, yet standard fusion mechanisms do not explicitly estimate whether a modality is reliable for the current prediction~\cite{ma2024sdt}. Third, discrete emotion classifiers generally treat emotion categories as independent labels, ignoring their structure in affective space and making emotionally shifted utterances particularly difficult to recognize, as in Fig.~\ref{fig:erc_example}.

This work addresses these limitations by extending the Self-Distillation Transformer (SDT)~\cite{ma2024sdt}, a transformer-based ERC architecture that jointly processes text, audio, and visual representations through intra- and inter-modal attention, hierarchical gated fusion, and unimodal self-distillation.
Building on this multimodal framework, we investigate whether stronger facial representations can make the visual stream more informative, whether modality contributions can be adapted to the predicted emotion class, and whether affective geometry can improve recognition under emotional transitions.

Specifically, this work proposes the following contributions:

\begin{itemize}
    \item A combination of visual encoders for appearance-based features with facial geometry descriptors to strengthen the visual stream;
    \item A class-wise adaptive modality fusion strategy that allows modality importance to vary across emotion categories;
    \item A valence-arousal prior that applies a shift-aware correction in the emotion space to improve recognition of emotionally shifted utterances;
\end{itemize}

The proposed method is evaluated on the MELD~\cite{poria2019meld} and IEMOCAP~\cite{busso2008iemocap} benchmarks for ERC. The experiments show consistent gains in multimodal fusion performance, together with targeted improvements in emotion-shift recognition that vary with the characteristics of each dataset. Code and pretrained weights are available at: \url{https://github.com/multimedia-eurecat/classwise-multimodal-ERC}.

%% file: sections/2_relatedwork.tex
\section{Related Work}
\label{sec:related_work}

\subsection{Multimodal Emotion Recognition in Conversations}

Emotion Recognition in Conversations differs from isolated emotion recognition
because predictions depend on conversational context, speaker identity, and the
temporal evolution of affect. Early methods used
recurrent or memory-based architectures to model these dependencies. CMN~\cite{hazarika2018cmn}
and ICON~\cite{hazarika2018icon} maintain conversational memories over
multimodal inputs, while DialogueRNN~\cite{majumder2019dialoguernn} explicitly
tracks speaker states, global context, and emotion dynamics. Graph-based
approaches such as DialogueGCN~\cite{ghosal2019dialoguegcn} and
MMGCN~\cite{hu2021mmgcn} instead represent utterances and speaker relations as
structured graphs.

More recent methods rely on attention-based and Transformer architectures to jointly
model long-range context and cross-modal interactions.
HiTrans~\cite{li2020hitrans} captures global and speaker-sensitive context,
DialogueTRM~\cite{mao2021dialoguetrm} combines hierarchical contextual modeling
with interactive multimodal fusion, and MM-DFN~\cite{hu2022mmdfn} integrates
graph reasoning with dynamic fusion. MMTr~\cite{zou2022mmtr} uses
cross-modal attention to strengthen weaker modalities, while
UniMSE~\cite{hu2022unimse} combines multi-level fusion with contrastive
alignment.

Transformer-based fusion is particularly suitable for multimodal dialogue
because it can model both intra-modal context and interactions between text,
audio, and vision. MulT~\cite{tsai2019mult} introduced cross-modal
attention for unaligned multimodal sequences, while
MAG~\cite{rahman2020mag} injects acoustic and visual information into a
pretrained language representation. However, standard attention and gating
mechanisms do not explicitly estimate whether a modality is informative or
reliable for a particular utterance.

\subsection{SDT Framework in a Nutshell}
\label{sec:related_sdt}

Our work builds on the Self-Distillation Transformer
(SDT)~\cite{ma2024sdt}, a multimodal ERC architecture that operates on utterance-level text, audio, and visual features. For each modality $m \in \{t, a, v\}$, the input features are first projected to a shared hidden dimension and enriched with positional and speaker embeddings. An intra-modal Transformer then models contextual dependencies within the same modality, while two inter-modal Transformers allow each modality to attend to the other two streams. The resulting self- and cross-modal representations are combined through unimodal-level sigmoid gates, producing one enhanced representation per modality, denoted as $\mathbf{H}'_m$.

At the multimodal level, SDT applies a softmax-based gate to combine the different modality enhanced representations $\mathbf{H}'_m$ before the final classification $\hat{\mathbf{Y}}$. In parallel, each unimodal branch has its own classifier and is trained through self-distillation, using the fused multimodal prediction $\hat{\mathbf{Y}}$ as a teacher of the unimodal-level classifiers. This encourages stronger unimodal representations while preserving a joint multimodal decision.

Our work retains SDT's intra- and inter-modal modality encoding and self-distillation strategy, but revisits the fusion stage by replacing the original softmax gate with class-wise adaptive modality fusion.

\subsection{Visual Representations for Emotion Recognition}

Visual ERC systems commonly rely on appearance-based encoders that extract a
holistic representation from face crops. Convolutional networks such as
DenseNet~\cite{huang2017densenet} and more recent Vision Transformers
(ViTs)~\cite{dosovitskiy2021vit} capture rich visual information, but may also
encode identity, illumination, pose, texture, and background together with
expression-related cues.

Geometry-based representations provide a complementary alternative by focusing
on facial structure and deformation. The Facial Action Coding System
(FACS)~\cite{ekman1978facs} represents expressions through anatomically
defined facial action units, while 3D Morphable Models
(3DMMs)~\cite{blanz1999morphable} separate identity-related facial structure
from expression parameters. Although appearance- and geometry-based features have been widely studied for facial expression recognition, their complementarity within multimodal ERC has received less attention~\cite{jung2015joint, zheng2023poster}.

\subsection{Emotion Dynamics and Affective Geometry}

Emotion shifts remain particularly challenging in ERC. Existing methods often
model them as a discrete auxiliary task. For example,
Bansal et al.~\cite{bansal2022shapes} predict whether an emotional change occurs
between consecutive utterances and use this signal to regulate the influence of
dialogue history on the next prediction.

A separate line of work represents relationships between emotions through
continuous affective dimensions. Russell's circumplex model
~\cite{russell1980circumplex} organizes  emotion categories around a continuous two-dimensional space according to valence and arousal, where distances between valence-arousal coordinates reflect affective similarity. Such distances have
been incorporated into training objectives to penalize confusions between
affectively distant classes more strongly~\cite{feng2023chatter}.

%% file: sections/3_method.tex
\section{Method}
\label{sec:method}

\begin{figure*}[t]
    \centering
    \includegraphics[width=0.99\linewidth]{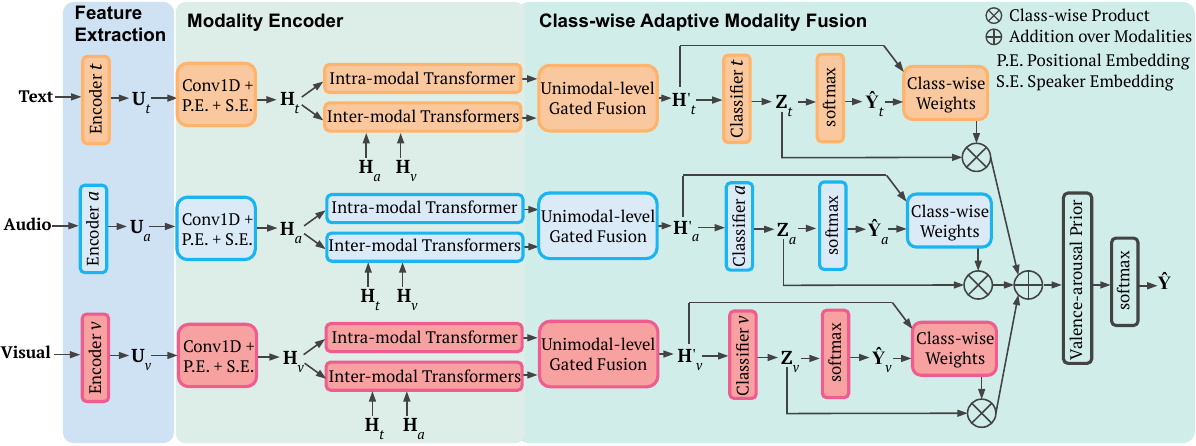}
    \caption{Method overview. Building on SDT~\cite{ma2024sdt}, unimodal features $\mathbf{U}$ are projected into $\mathbf{H}$ and processed by intra- and inter-modal Transformers to obtain enhanced representations $\mathbf{H}'$. We introduce class-wise adaptive modality fusion and a valence-arousal prior that biases the final prediction $\hat{\mathbf{Y}}$ toward plausible affective transitions.}
    \label{fig:diagram}
\end{figure*}

This section presents the proposed methodology for ERC from text, audio, and visual streams. As shown in Fig.~\ref{fig:diagram}, the pipeline comprises three main stages: unimodal feature extraction, modality encoding, and class-wise adaptive modality fusion followed by a valence-arousal prior.

The unimodal feature extraction and modality encoding stages build on the Self-Distillation Transformer (SDT)~\cite{ma2024sdt}. We update the unimodal encoders and extend the visual stream by combining appearance-based and geometry-based facial cues. We then replace SDT's original softmax-based multimodal fusion with a class-wise adaptive strategy that combines unimodal logits using explicit estimates of each modality's informativeness for each emotion class. Finally, we introduce a valence-arousal prior grounded in affective geometry, which applies a targeted correction to the fused logits when the predicted affective state changes strongly across consecutive utterances.

The training objective follows the original SDT formulation:
\begin{equation}
    \mathcal{L} =
        \gamma_1 \, \mathcal{L}_{\mathrm{Task}}
      + \gamma_2 \, \mathcal{L}_{\mathrm{CE}}
      + \gamma_3 \, \mathcal{L}_{\mathrm{KL}},
    \label{eq:loss_function}
\end{equation}
where $\mathcal{L}_{\mathrm{Task}}$ and $\mathcal{L}_{\mathrm{CE}}$ are cross-entropy losses supervising the fused multimodal and modality-specific predictions, respectively, using the ground-truth labels. Self-distillation is applied via the Kullback-Leibler (KL) divergence term $\mathcal{L}_{\mathrm{KL}}$, which encourages the temperature-softened unimodal-level predictions to match the softened fused prediction; larger temperature values yield a softer distribution over classes~\cite{hinton2015distilling}.

\subsection{Feature Extraction and Modality Encoding}
\label{subsec:feature_extraction}
We replace the original SDT feature extractors~\cite{ma2024sdt} with more recent pretrained encoders to extract unimodal feature representations $\textbf{U}_m$ for each modality $m$. Following SDT, these representations are projected to a shared dimensionality $d=1024$ using $1\times1$ convolutions and combined with positional embeddings $\mathbf{PE}$ and speaker embeddings $\mathbf{SE}$, \ie, $\textbf{H}_m = \text{Conv1D}(\textbf{U}_m) + \textbf{PE} + \textbf{SE}$.
The modality encoding stage, shown in Fig.~\ref{fig:diagram}, remains unchanged and uses intra-modal and inter-modal Transformers to capture contextual relationships within each modality and interactions across modalities.

\paragraph{Text features.}
Text embeddings are extracted with RoBERTa-large trained within the sentence-transformer framework~\cite{reimers2019sentencebert}\footnote{\scriptsize \url{https://huggingface.co/sentence-transformers/all-roberta-large-v1}}. Each target utterance is preceded by the two previous dialogue turns for context, formatted with speaker identifiers, and encoded by mean-pooling the final hidden states.

\paragraph{Audio features.}
Audio embeddings are extracted with the speech emotion representation model emotion2vec~\cite{ma2023emotion2vec}\footnote{\scriptsize \url{https://huggingface.co/emotion2vec/emotion2vec_plus_large}}, trained on large-scale, unspecified speech emotion recognition corpora, replacing the handcrafted openSMILE features~\cite{eyben2013opensmile} used by SDT. As the exact training subsets are not publicly disclosed, possible overlap with IEMOCAP or MELD cannot be determined. The publicly released checkpoint is used here without fine-tuning. Audio is converted to mono, resampled to 16\,kHz, RMS-normalized, and truncated to 10\,s. Missing or invalid segments are represented by zero vectors.

\begin{figure}[t]
  \centering
  \includegraphics[width=\columnwidth]{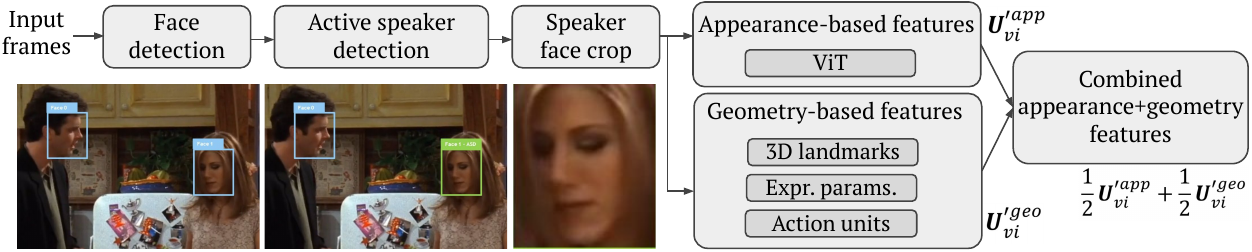}
  \caption{
  Overview of the visual feature extraction. Speaker face crops are obtained from $T=16$ frames per utterance, then encoded as a combined representation of appearance- and geometric-based facial features before entering the visual modality encoding stage.
  }
  \label{fig:visual_encoders}
\end{figure}

\paragraph{Visual features.}
For each utterance, $T=16$ frames are sampled uniformly. Faces are detected with MTCNN~\cite{zhang2016mtcnn}, while LightASD~\cite{liao2023lightasd} selects the active speaker in multi-person scenes. If no valid face is found, the visual embedding is set to zero. The resulting face crops are encoded using a combination of appearance-based features and geometry-based descriptors, as illustrated in Fig.~\ref{fig:visual_encoders}.

Appearance features are obtained with a Vision Transformer (ViT)~\cite{dosovitskiy2021vit}\footnote{ \scriptsize \url{https://huggingface.co/dima806/facial_emotions_image_detection}}. Each frame is encoded independently, and the resulting 768-dimensional embeddings are mean-pooled into one utterance-level descriptor.

To provide a more structured representation of facial expressions, we evaluate three geometry-based descriptors:
\begin{enumerate}
    \item \textit{3D landmarks}: 68 landmarks extracted with 3DDFA-V2~\cite{guo2020towards}, normalized by nose position and inter-ocular distance, and summarized across frames using their mean and standard deviation, producing a 408-dimensional descriptor;

    \item \textit{Expression parameters}: 10 expression coefficients and three head-pose angles estimated with 3DDFA-V2 and a 3D Morphable Model~\cite{blanz1999morphable}, aggregated using the mean, standard deviation, and average frame-to-frame difference into a 39-dimensional descriptor;

    \item \textit{Action units}: 20 facial action-unit intensities defined by FACS~\cite{ekman1978facs} and extracted with py-feat~\cite{cheong2023pyfeat}, aggregated using the same three statistics into a 60-dimensional descriptor.
\end{enumerate}
Appearance-based descriptors are projected to the shared dimension using a
$1\times1$ convolution as in SDT, while geometry-based descriptors are mapped to the same
space using a lightweight two-layer MLP that can model non-linear relationships within
the structured facial features. After independent normalization, the two
branches are combined by symmetric addition: 
$
    \mathbf{U}'_{vi}
    =
    \frac{1}{2}\mathbf{U}'^{\mathrm{app}}_{vi}
    +
    \frac{1}{2}\mathbf{U}'^{\mathrm{geo}}_{vi},
    \label{eq:combined_visual}
$
where
$\mathbf{U}'^{\mathrm{app}}_{vi},
 \mathbf{U}'^{\mathrm{geo}}_{vi}\in\mathbb{R}^{d}$
are the projected and normalized appearance and geometry embeddings for
utterance $i$, and $\mathbf{U}'_{vi}$ is the combined visual representation previous to the addition of the positional and speaker embeddings.

\subsection{Class-Wise Adaptive Modality Fusion}
\label{subsec:fusion_strategies}
The original SDT architecture uses hierarchical gated fusion at both the unimodal and multimodal levels~\cite{ma2024sdt}. At the unimodal level, the outputs of the intra- and inter-modal Transformers are combined into an enhanced representation $\mathbf{H}'_m$ for each modality, as shown in Fig.~\ref{fig:diagram}. At the multimodal level, these representations are fused through a softmax-based gate. 

This multimodal gate is not explicitly reliability-aware: a modality may receive a large contribution because of the magnitude of its projected activations, even when its signal is noisy, corrupted, or weakly informative. We therefore retain
the unimodal-level gated fusion and replace the softmax-based gate with a class-wise adaptive fusion strategy that estimates the contribution of each modality separately for each emotion category.

Throughout this section, $\mathbf{h}'_{mi} \in \mathbb{R}^{d}$ denotes the $i$-th row of $\mathbf{H}'_m$, \ie the enhanced representation of modality $m$ for a given utterance. Each modality-specific classifier maps $\mathbf{h}'_{mi}$ to unimodal logits $\mathbf{z}_{mi} \in \mathbb{R}^{C}$ in $\mathbf{Z}_m$ and a subsequent probability vector $\hat{\mathbf{y}}_{mi} = \mathrm{softmax}(\mathbf{z}_{mi}) \in \mathbb{R}^C$ in $\hat{\mathbf{Y}}_m$ over the $C$ emotion categories.

\paragraph{Multimodal softmax fusion.}
The SDT baseline~\cite{ma2024sdt} applies a shared linear projection $\mathbf{W} \in \mathbb{R}^{d \times d}$ to each enhanced modality representation $\mathbf{h}'_{mi}$, where $m \in \{t,a,v\}$ denotes text, audio, or visual stream. A softmax over the modality dimension produces dimension-wise weights as follows:
\begin{equation}
    \left[\mathbf{g}_{ti};\, \mathbf{g}_{ai};\, \mathbf{g}_{vi}\right]
    = \text{softmax}\!\left(
        \left[\mathbf{W}\mathbf{h}'_{ti};\,
              \mathbf{W}\mathbf{h}'_{ai};\,
              \mathbf{W}\mathbf{h}'_{vi}\right]
    \right).
    \label{eq:softmax_org}
\end{equation}
The final multimodal representation $\mathbf{h}'_i$ is obtained as a weighted sum of the enhanced modality representations:
\begin{equation}
    \mathbf{h}'_i = \sum_{m \in \{t,a,v\}}
                    \mathbf{g}_{mi} \odot \mathbf{h}'_{mi},
\end{equation}
where $\odot$ denotes the element wise-product. Although expressive, this gate does not explicitly estimate whether each modality is informative for the current utterance and a certain emotion class.

\paragraph{Multimodal class-wise adaptive fusion.}
The relevance of a modality may depend on the emotion being predicted. Audio, for example, may be especially informative for high-arousal emotions, whereas visual cues may be more useful for classes associated with distinctive facial configurations. Our class-wise adaptive fusion implements this idea by estimating an explicit contribution score for each emotion class $c$ and modality $m$, based on $\mathbf{h}'_{mi} \in \mathbb{R}^{d}$ and $\hat{\mathbf{y}}_{mi} \in \mathbb{R}^{C}$, as follows:
\begin{equation}
    r_{mi}^{c} = \mathcal{R}_{m}^{c}\!\left([\mathbf{h}'_{mi} \,\|\, \hat{\mathbf{y}}_{mi}]\right),
    \label{eq:classwise_reliability}
\end{equation}
where $\|$ denotes concatenation and $\mathcal{R}_{m}$ is a learnable function per modality implemented as a lightweight two-layer MLP. Its $c$-th output, $r_{mi}^{c}$, represents the estimated informativeness of the modality $m$ for the emotion class $c$.

For each class, $r_{mi}^{c}$ scores in Eq.~\eqref{eq:classwise_reliability} are converted to weights via softmax across modalities, $w_{mi}^{c} = \text{softmax}_m(r_{mi}^{c})$. Note that fusion is performed at the logit level
rather than directly on $\mathbf{h}'_{mi}$. The fused logit for class $c$ is obtained as
\begin{equation}
    \hat{z}_{i}^{c} = \sum_{m \in \{t,a,v\}} w_{mi}^{c} \, z_{mi}^{c} ,
\end{equation}
where $\hat{z}_{i}^{c}$ and $z_{mi}^{c}$ are the $c$-class values of the fused and unimodal-level logits, $\hat{\mathbf{z}}_{mi}$ and $\mathbf{z}_{mi}$, respectively.
The final prediction is obtained from the fused logits as $\hat{\mathbf{y}}_i = \text{softmax}(\hat{\mathbf{z}}_i)$. The unimodal predictions provided to the class-wise learnable MLPs are detached from the computation graph, preventing the fusion weights from influencing the unimodal classifiers through a circular gradient path.

\subsection{Valence-Arousal Prior for Emotion Shifts}
\label{subsec:circumplex_grounded_prior}

\begin{figure}[t]
  \centering
  \includegraphics[width=\columnwidth]{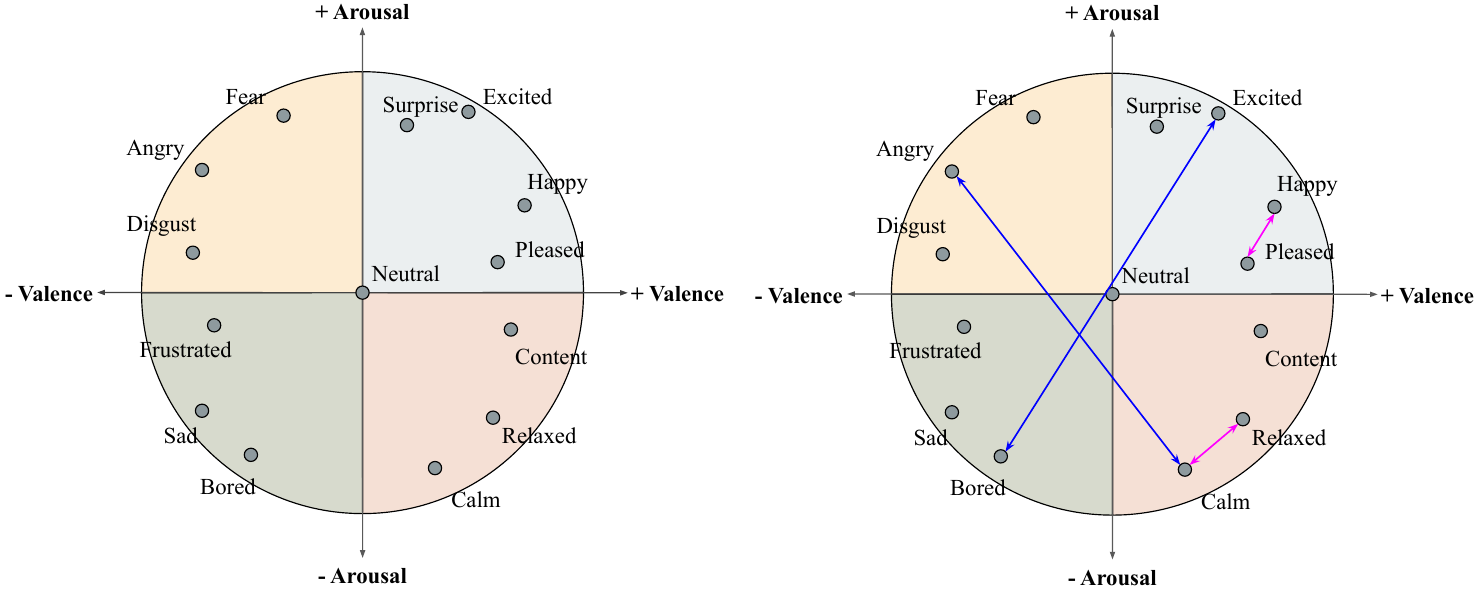}
  \caption{
  Russell’s valence-arousal circumplex model of affect. The proposed prior favors smaller, plausible emotion transitions (magenta) over large shifts (blue).
  }
  \label{fig:circumplex}
\end{figure}

Emotion shifts remain challenging in ERC because standard classifiers treat emotion categories as independent labels, ignoring their relationships in affective space. For example, a transition from \textit{happy} to \textit{excited} is affectively closer than one from \textit{happy} to \textit{sad}. We therefore introduce a shift-aware prior based on Russell's circumplex model~\cite{russell1980circumplex}, which represents emotions in a two-dimensional continuous space defined by valence and arousal, as illustrated in Fig.~\ref{fig:circumplex}.
In this space, each emotion class $c$ is assigned a pair of valence-arousal coordinates
$\mathbf{v}_c \in \mathbb{R}^{2}$.
These coordinates can be defined using either canonical circumplex positions or dataset-specific centroids derived from continuous annotations.

We apply a valence-arousal prior after class-wise adaptive fusion, which produces multimodal logits
$\hat{\mathbf{z}}_i \in \mathbb{R}^{C}$ and corresponding class probabilities
$
    \hat{\mathbf{y}}_i
    =
    \mathrm{softmax}(\hat{\mathbf{z}}_i).
$
The expected valence-arousal position of utterance $i$ is then computed as
\begin{equation}
    \hat{\mathbf{v}}_i
    =
    \sum_{c=1}^{C}
    \hat{y}_i^{c}\mathbf{v}_c,
\end{equation}
where $\hat{y}_i^{c}$ is the predicted probability of class $c$. The same estimate is obtained for the preceding utterance $\hat{\mathbf{v}}_{i-1}$. For the first utterance of a dialogue, the previous class distribution is initialized uniformly.

The magnitude of the predicted affective transition is measured by the Euclidean distance between two consecutive valence-arousal positions:
\begin{equation}
    \beta_i
    =
    \left\|
        \hat{\mathbf{v}}_i
        -
        \hat{\mathbf{v}}_{i-1}
    \right\|,
    \label{eq:velocity}
\end{equation}
so that the correction has little influence on affectively stable utterances and becomes stronger for larger predicted shifts.

For each candidate class $c$, the distance from the previous affective state is
\begin{equation}
    d_i^{c}
    =
    \left\|
        \mathbf{v}_c
        -
        \hat{\mathbf{v}}_{i-1}
    \right\|.
\end{equation}
These distances are converted into a temperature-scaled log-prior as
\begin{equation}
    \phi_i^{c}
    =
    \log
    \frac{
        \exp(-d_i^{c}/\tau_{\mathrm{va}})
    }{
        \sum_{c'=1}^{C}
        \exp(-d_i^{c'}/\tau_{\mathrm{va}})
    },
    \label{eq:va_prior_temp}
\end{equation}
where smaller $\tau_{\mathrm{va}}$ assigns greater preference to classes closer to the previous affective state. Stacking the class-wise values gives the log-prior vector
$\boldsymbol{\phi}_i \in \mathbb{R}^{C}$, which is added to the fused logits:
\begin{equation}
    \hat{\mathbf{z}}'_i
    =
    \hat{\mathbf{z}}_i
    +
    \alpha \beta_i \boldsymbol{\phi}_i,
    \label{eq:va_prior_final}
\end{equation}
where $\alpha \geq 0$ controls the overall prior strength and $\beta_i$ scales the correction according to the predicted transition magnitude. In this way, the prior acts as an affective transition smoother: $\beta_i$ determines how strongly the correction is applied, while $\boldsymbol{\phi}_i$ favors emotion classes that remain close to the previous affective state. The final prediction is $\hat{\mathbf{y}}'_i=\mathrm{softmax}(\hat{\mathbf{z}}'_i).$ The prior correction term is applied at both training and inference; $\alpha$ and $\tau_{\mathrm{va}}$ are fixed by hyperparameter sweep and no gradients flow through Eqs.~\eqref{eq:velocity}--\eqref{eq:va_prior_final}. The prior correction at utterance $i$ uses the base class probabilities of the previous utterance, before adding the prior correction, so that errors are not propagated along a dialogue.

%% file: sections/4_experiments.tex
\section{Experiments}
\label{sec:experiments}

\definecolor{rankone}{HTML}{F4A6A6}
\definecolor{ranktwo}{HTML}{FBD9A8}

This section evaluates the proposed components on ERC datasets through a progressive ablation. After presenting the common experimental setup and overall comparison, we separately analyze the appearance+geometry visual representations, class-wise adaptive modality fusion, and the
valence-arousal prior for emotionally shifted utterances.

\subsection{Experimental Setup}
\label{subsec:experimental_setup}

\paragraph{Datasets.}
We evaluate the proposed method on two multimodal ERC benchmarks: MELD~\cite{poria2019meld} and IEMOCAP~\cite{busso2008iemocap}. Both provide utterance-level emotion labels together with text, audio, and video, but differ substantially in conversational setting and annotation structure.

MELD contains 13,708 utterances from 1,433 multi-party dialogues, annotated with seven emotion classes: \textit{neutral}, \textit{surprise}, \textit{fear},
\textit{sadness}, \textit{joy}, \textit{disgust}, and \textit{anger}. We use the official training, validation, and test splits, containing 9,989, 1,109, and 2,610 utterances, respectively. IEMOCAP contains 7,433 utterances from dyadic interactions and uses six classes: \textit{happy}, \textit{sad}, \textit{neutral}, \textit{angry}, \textit{excited}, and \textit{frustrated}. Following the SDT protocol~\cite{ma2024sdt}, Sessions 1-4 are used for training and Session 5 for testing. IEMOCAP additionally provides continuous valence and arousal annotations, which are used only in the emotion-shift analysis.

\paragraph{Evaluation protocol.}
Weighted F1 is used as the primary metric because both datasets exhibit
substantial class imbalance; overall accuracy is reported as a complementary
measure. For the valence-arousal prior, we additionally report accuracy
separately on emotionally shifted and stable utterances.
For MELD, the official validation and test splits are used.
IEMOCAP does not have a validation split; following the SDT protocol~\cite{ma2024sdt}, Session 5 is used for both validation and test. 
Hyperparameter choices are made using validation data, and results are reported on the test set.

\paragraph{Baseline and training.}
All experiments build on our reimplementation of SDT~\cite{ma2024sdt}, using
the updated text, audio, and visual encoders described in
Section~\ref{subsec:feature_extraction}. The original SDT results are included
only as reported reference values because its fine-tuned feature-extractor
weights were not publicly available. Unless otherwise stated, ablations modify
one component at a time while keeping the remaining architecture and training
configuration fixed.

As in SDT, for MELD, we use a learning
rate of $5\times10^{-6}$, batch size 8, and self-distillation temperature
$\tau=8$; and for IEMOCAP, we use a learning rate of $10^{-4}$, batch size 16, and
$\tau=1$. Models are trained for up to 15 epochs with early stopping based on weighted F1 of the validation set. All configurations used a fixed random seed to ensure consistent comparisons across ablations. The SDT softmax fusion gate requires a $\sim$1.05M projection layer, whereas our class-wise approaches requires only $\sim$0.40M. All experiments were run on a single NVIDIA GTX 1080 Ti (12\,GB VRAM).

\begin{table*}[t]
\centering
\caption{Comparison with established multimodal ERC methods and cumulative
ablation of the proposed components. Results report overall accuracy (ACC)
and weighted F1 (w-F1), in \%. Mean denotes the average
across MELD and IEMOCAP. Results marked with $\dagger$ are taken
from~\cite{ma2024sdt} and were not reproduced in this work.}
\label{tab:overall_ablation}
\renewcommand{\arraystretch}{1.15}
\setlength{\tabcolsep}{5pt}
\resizebox{0.9\textwidth}{!}{%
\begin{tabular}{l cc cc cc}
\toprule
& \multicolumn{2}{c}{\textbf{MELD}}
& \multicolumn{2}{c}{\textbf{IEMOCAP}}
& \multicolumn{2}{c}{\textbf{Mean}} \\
\cmidrule(lr){2-3}
\cmidrule(lr){4-5}
\cmidrule(lr){6-7}
\textbf{Model}
& \textbf{ACC} & \textbf{w-F1}
& \textbf{ACC} & \textbf{w-F1}
& \textbf{ACC} & \textbf{w-F1} \\
\midrule

\multicolumn{7}{l}{\textit{Previously reported ERC methods}} \\[1pt]

DialogueRNN~\cite{majumder2019dialoguernn}$^{\dagger}$
& 66.70 & 65.31
& 69.38 & 69.37
& 68.04 & 67.34 \\

MMGCN~\cite{hu2021mmgcn}$^{\dagger}$
& 66.40 & 65.21
& 69.62 & 69.61
& 68.01 & 67.41 \\

DialogueTRM~\cite{mao2021dialoguetrm}$^{\dagger}$
& 66.70 & 65.76
& 69.87 & 69.93
& 68.29 & 67.85 \\

MM-DFN~\cite{hu2022mmdfn}$^{\dagger}$
& 66.55 & 65.48
& 69.87 & 69.91
& 68.21 & 67.70 \\

Original SDT~\cite{ma2024sdt}$^{\dagger}$
& 67.55 & 66.60
& 73.95 & 74.08
& 70.75 & 70.34 \\

\midrule
\multicolumn{7}{l}{\textit{Cumulative ablation of our model}} \\[1pt]

Updated SDT baseline
& 74.87 & 75.49
& 69.36 & 69.50
& 72.12 & 72.50 \\

+ Appearance+geometry visual stream
& \textbf{75.40} & 75.76
& 73.98 & 73.86
& 74.69 & 74.81 \\

+ Class-wise adaptive modality fusion
& 75.36 & \textbf{75.93}
& \textbf{74.72} & \textbf{74.11}
& \textbf{75.04} & \textbf{75.02} \\

\bottomrule
\end{tabular}%
}
\end{table*}

\subsection{Overall Comparison and Ablation}
\label{subsec:overall_ablation}

We first evaluate the cumulative effect of the proposed components on the multimodal text-audio-visual prediction, starting from our updated SDT baseline and progressively adding the proposed appearance+geometry visual representation and class-wise adaptive modality fusion.

Table~\ref{tab:overall_ablation} places the proposed method in the context of
established multimodal ERC approaches and reports the cumulative effect of the
visual and fusion components. Previously published results are included as
reference values and were not reproduced under our updated feature-extraction
setting.

The updated feature encoders affect the two datasets differently, substantially
improving MELD while reducing performance relative to the reported SDT result
on IEMOCAP. Nevertheless, their mean weighted F1 across both datasets increases
from 70.34 to 72.50. Adding appearance+geometry visual representations raises
the mean weighted F1 to 74.81, with particularly strong gains on IEMOCAP.
Class-wise adaptive modality fusion provides a further improvement to 75.02
mean weighted F1 and achieves the strongest IEMOCAP accuracy and weighted F1.
Sections~\ref{subsec:visual_analysis} and~\ref{subsec:fusion_analysis} analyze
these contributions individually.
The valence-arousal prior is separately examined in  Section~\ref{subsec:emotion_shift_results}.

\subsection{Visual Representation Analysis}
\label{subsec:visual_analysis}

This section analyzes the effect of appearance- and geometry-based facial representations on multimodal ERC performance, as summarized in Table~\ref{tab:visual_encoders}. To isolate the visual stream, all configurations retain the updated text and audio encoders and use the original SDT softmax fusion. We compare ViT appearance features with 3D landmarks, expression parameters, and facial action units, both individually and in combination with ViT according to Fig.~\ref{fig:visual_encoders}. All geometry-based descriptors are projected to the shared hidden space using the lightweight MLP described in Section~\ref{subsec:feature_extraction}.

\begin{table}[t]
\centering
\caption{Visual representation ablation on MELD and IEMOCAP, reported as
trimodal overall accuracy (ACC) and weighted F1 (w-F1), in \%.
Best and second-best results are shown in bold and underlined, respectively.}
\label{tab:visual_encoders}
\small
\setlength{\tabcolsep}{5pt}
\renewcommand{\arraystretch}{1.15}

\resizebox{0.65\linewidth}{!}{%
\begin{tabular}{l cc cc}
\toprule
& \multicolumn{2}{c}{\textbf{MELD}}
& \multicolumn{2}{c}{\textbf{IEMOCAP}} \\
\cmidrule(lr){2-3}
\cmidrule(lr){4-5}
\textbf{Visual representation}
& \textbf{ACC} & \textbf{w-F1}
& \textbf{ACC} & \textbf{w-F1} \\
\midrule
ViT appearance
& 74.87 & 75.49
& 69.36 & 69.50 \\
\midrule
3D landmarks
& 74.98 & 75.34
& 72.93 & 72.41 \\

Expression parameters
& 75.21 & \underline{75.59}
& 73.00 & 72.51 \\

Action units
& 75.17 & 75.43
& \underline{73.30} & \underline{73.20} \\
\midrule
ViT + 3D landmarks
& \textbf{75.59} & 75.27
& \textbf{73.98} & \textbf{73.86} \\

ViT + expression parameters
& 74.79 & 75.47
& 70.59 & 70.06 \\

ViT + action units
& \underline{75.40} & \textbf{75.76}
& 71.52 & 71.67 \\
\bottomrule
\end{tabular}%
}
\end{table}

On MELD, geometry-only representations perform similarly to the ViT appearance baseline, indicating that structured facial descriptors retain useful expression information despite their substantially lower dimensionality. Expression parameters provide the strongest geometry-only result, while combining ViT with action units achieves the best overall performance.

The effect of facial geometry is more pronounced on IEMOCAP, where all three geometry-only representations outperform ViT appearance features. Action units provide the strongest geometry-only result compared with ViT, while the best overall configuration combines ViT with 3D landmarks.

The results also show that combining appearance and geometry is not uniformly beneficial. While action units complement ViT on MELD and 3D landmarks complement it on IEMOCAP, the remaining combinations perform below their corresponding geometry-only representations. This indicates that the usefulness of appearance+geometry fusion depends on both the facial descriptor and the dataset characteristics. Based on the weighted F1 results in Table~\ref{tab:visual_encoders}, subsequent experiments use ViT with action units for MELD and ViT with 3D landmarks for IEMOCAP. 

\subsection{Class-Wise Adaptive Fusion Analysis}
\label{subsec:fusion_analysis}

We evaluate whether estimating modality informativeness separately for each emotion
class improves over class-independent fusion. All configurations use the visual
representations selected in Section~\ref{subsec:visual_analysis}: ViT combined
with action units for MELD and with 3D landmarks for IEMOCAP. We compare the
original SDT softmax gate, the proposed class-wise strategy and a modality-wise but class-agnostic variant. For the modality-wise variant, the learnable MLP in Eq.~$\eqref{eq:classwise_reliability}$ is modified to predict a single score per modality instead of $C$ class-specific scores.

As reported in Table~\ref{tab:overall_ablation}, class-wise adaptive fusion improves
weighted F1 from 75.76 to 75.93 on MELD and from 73.86 to 74.11 on IEMOCAP.
In contrast, the modality-wise variant obtains only 73.47
and 73.17 weighted F1 on MELD and IEMOCAP, respectively. 
A single score per modality therefore appears too
coarse to represent the emotion-dependent informativeness of text, audio, and visual
cues. 
On IEMOCAP, class-wise
fusion increases accuracy, while 
on MELD it remains
essentially unchanged.
This suggests that the class-wise weights mainly improve the balance
of predictions across MELD's imbalanced emotion categories rather than the
total number of correct predictions. The larger effect on IEMOCAP is also
consistent with the stronger contribution of facial geometry observed in
Section~\ref{subsec:visual_analysis}, leaving more room for class-dependent
multimodal weighting.

\begin{table}[t]
\centering
\caption{Per-class weighted F1 on MELD and IEMOCAP for the SDT softmax baseline and the proposed class-wise adaptive fusion, with the visual encoders selected in Sec.~\ref{subsec:visual_analysis}
}
\label{tab:per_class_f1}
\footnotesize
\setlength{\tabcolsep}{9pt}
\renewcommand{\arraystretch}{0.95}
\resizebox{0.85\linewidth}{!}{%
\begin{tabular}{lcc@{\hskip 1em}lcc}
\toprule
\multicolumn{3}{c}{\textbf{MELD}} & \multicolumn{3}{c}{\textbf{IEMOCAP}} \\
\cmidrule(lr){1-3} \cmidrule(lr){4-6}
Class & Softmax & Class-wise & Class & Softmax & Class-wise \\
\midrule
Neutral   & 96.16          & \textbf{96.40} & Happy      & \textbf{57.34} & 56.81          \\
Surprise  & 45.19          & \textbf{46.15} & Sad        & 80.82          & \textbf{81.84} \\
Fear      & 10.08          & \textbf{15.29} & Neutral    & 79.44          & \textbf{80.94} \\
Sadness   & 51.24          & \textbf{52.96} & Angry      & 72.46          & \textbf{74.24} \\
Joy       & 77.60          & \textbf{77.76} & Excited    & 80.13          & \textbf{81.13} \\
Disgust   & \textbf{31.40} & 27.33          & Frustrated & \textbf{65.66} & 63.17          \\
Anger     & \textbf{57.27} & 55.78          &            &                &                \\
\midrule
Average & 75.76 & \textbf{75.93} & Average & 73.86 & \textbf{74.11} \\
\bottomrule
\end{tabular}%
}
\end{table}

Table~\ref{tab:per_class_f1} reports the per-class weighted F1 for both fusion strategies. Class-wise gains concentrate on less-represented emotions on both datasets, while softmax fusion remains stronger on a small subset of classes. In general, no clear pattern is observed between positive and negative emotions.

We further trained on partial missing modalities with noisy partitions: 50\% clean, $\sim$50\% audio-, visual-, or both-degraded. Table~\ref{tab:degradation_robustness} shows that MELD, dominated by text, does not benefit from degradation training. However, IEMOCAP's clean-trained model collapses under audio degradation and drops sharply under visual degradation; degradation training recovers most of this loss.
While MELD is sourced from television dialogues, with natural environmental noise, varied lighting and occasionally missing speaker faces,
IEMOCAP was recorded in a controlled laboratory setting. This explains why introducing degraded data in training benefits models trained on IEMOCAP by improving their robustness.

\begin{table}[t]
\centering
\caption{Weighted F1 under clean and degraded test conditions for clean-trained and degradation-trained models. Audio degraded averages SNR\,=\,5\,dB and 50\% packet loss; visual degraded averages $\sigma\!=\!7$ blur and 50\% lower-face occlusion.}
\label{tab:degradation_robustness}
\small
\setlength{\tabcolsep}{8pt}
\renewcommand{\arraystretch}{1.12}
\resizebox{0.85\linewidth}{!}{%
\begin{tabular}{lcccc}
\toprule
 & \multicolumn{2}{c}{\textbf{MELD}} & \multicolumn{2}{c}{\textbf{IEMOCAP}} \\
\cmidrule(lr){2-3}\cmidrule(lr){4-5}
\textbf{Test condition}
  & \textbf{No degraded} & \textbf{Degraded}
  & \textbf{No degraded} & \textbf{Degraded} \\
\midrule
Clean          & \textbf{75.93} & 75.61 & \textbf{74.11} & 68.81 \\
Audio degraded    & \textbf{72.48} & 71.31 & 20.61 & \textbf{47.79} \\
Visual degraded   & \textbf{75.63} & 73.96 & 56.66 & \textbf{64.97} \\
\bottomrule
\end{tabular}%
}
\end{table}

\subsection{Emotion-Shift Analysis}
\label{subsec:emotion_shift_results}

We evaluate whether the valence-arousal prior introduced in
Section~\ref{subsec:circumplex_grounded_prior} improves recognition of
utterances involving emotional transitions. An utterance is
considered \emph{shifted} when its ground-truth emotion differs from that of
the most recent preceding utterance by the same speaker, and \emph{stable}
otherwise. Utterances without a preceding turn from the same speaker are
excluded. This yields 1,003 shifted and 861 stable utterances for MELD, and
410 shifted and 1,151 stable utterances for IEMOCAP. The prior remains dialogue-level and uses the immediately preceding dialogue turn when estimating
the previous affective state.

For this experiment, the configurations with and without the valence--arousal
prior both use class-wise adaptive fusion and the visual representations selected
in Section~\ref{subsec:visual_analysis}: ViT combined with action units for MELD
and with 3D landmarks for IEMOCAP. Both configurations also use input-level
modality dropout with $p_{\mathrm{drop}}=0.10$ as a fixed training regularizer,
randomly zeroing modality inputs while ensuring that at least one stream remains
available.

The two datasets provide different sources to represent the affective space.
IEMOCAP class coordinates are obtained by averaging the continuous
valence and arousal annotations associated with each emotion category. MELD
does not provide continuous affective annotations, so its classes are assigned
canonical coordinates derived from Russell's circumplex
model~\cite{russell1980circumplex}. All coordinates are normalized to
$[0,1]$ and shown in Table~\ref{tab:va_coordinates}. The prior parameters
are fixed to $\alpha=0.1$ and $\tau_{\mathrm{va}}=5$ for MELD, and
$\alpha=0.7$ and $\tau_{\mathrm{va}}=1.5$ for IEMOCAP.

\begin{table*}[t]
\centering
\caption{Valence-arousal coordinates normalized between $[0,1]$. IEMOCAP coordinates are class centroids derived from
continuous annotations, while MELD uses canonical positions based on
Russell's circumplex model~\cite{russell1980circumplex}.}
\label{tab:va_coordinates}
\small
\setlength{\tabcolsep}{5pt}
\renewcommand{\arraystretch}{1.12}

\resizebox{\textwidth}{!}{%
\begin{tabular}{llccccccc}
\toprule
\textbf{Dataset}
& \textbf{Dimension}
& \multicolumn{7}{c}{\textbf{Emotion class}} \\
\midrule

\multirow{3}{*}{\textbf{IEMOCAP}}
& \textbf{Emotion}
& Neutral & Happy & Excited & Angry & Sad & Frustrated & -- \\
& \textbf{Valence}
& 0.50 & 0.80 & 0.75 & 0.20 & 0.20 & 0.25 & -- \\
& \textbf{Arousal}
& 0.50 & 0.60 & 0.80 & 0.80 & 0.30 & 0.55 & -- \\
\midrule

\multirow{3}{*}{\textbf{MELD}}
& \textbf{Emotion}
& Neutral & Joy & Surprise & Anger & Sadness & Fear & Disgust \\
& \textbf{Valence}
& 0.50 & 0.88 & 0.70 & 0.29 & 0.19 & 0.18 & 0.20 \\
& \textbf{Arousal}
& 0.50 & 0.74 & 0.84 & 0.84 & 0.37 & 0.80 & 0.68 \\
\bottomrule
\end{tabular}%
}
\end{table*}

\begin{table}[t]
\centering
\caption{Effect of the valence-arousal prior on overall weighted F1 and
accuracy on emotionally shifted and stable utterances, in \%. Parentheses indicate the number of test utterances in each subset. Both
configurations use modality dropout with
$p_{\mathrm{drop}}=0.10$.}
\label{tab:emotion_shift}
\small
\setlength{\tabcolsep}{5pt}
\renewcommand{\arraystretch}{1.15}

\resizebox{0.9\linewidth}{!}{%
\begin{tabular}{l ccc ccc}
\toprule
& \multicolumn{3}{c}{\textbf{MELD}}
& \multicolumn{3}{c}{\textbf{IEMOCAP}} \\
\cmidrule(lr){2-4}
\cmidrule(lr){5-7}
\textbf{Configuration}
& \shortstack{\textbf{w-F1}\\$(1864)$}
& \shortstack{\textbf{ACC Shift}\\$(1003)$}
& \shortstack{\textbf{ACC Stable}\\$(861)$}
& \shortstack{\textbf{w-F1}\\$(1561)$}
& \shortstack{\textbf{ACC Shift}\\$(410)$}
& \shortstack{\textbf{ACC Stable}\\$(1151)$} \\
\midrule
Without prior
& 75.47
& 62.81
& 85.95
& 74.69
& 54.90
& \textbf{81.60} \\

With prior
& \textbf{75.61}
& \textbf{63.11}
& \textbf{86.06}
& \textbf{74.87}
& \textbf{55.64}
& \textbf{81.60} \\
\bottomrule
\end{tabular}%
}
\end{table}

As shown in Table~\ref{tab:emotion_shift}, the prior produces modest
improvements in overall weighted F1. 
This limited overall effect is consistent with its
targeted design: the velocity term reduces the correction for affectively stable
predictions and increases its influence when the predicted transition is larger.

The effect is more apparent on emotionally shifted utterances. 
The stronger improvement on IEMOCAP is consistent with
its dataset-specific valence-arousal coordinates, which are better aligned with
the empirical affective distribution than the canonical coordinates used for
MELD. 
Overall, these results are consistent with the intended role of the prior as an
affective transition smoother. It favors classes that remain plausible relative
to the preceding dialogue state while preserving performance on stable
utterances.

\subsection{Limitations}
\label{subsec:limitations}

The results presented in this work remain dataset dependent. The visual stream
relies on face detection and active-speaker identification, which may fail under
occlusion, profile views or overlapping speech.
The affective prior is also limited by the availability of dataset-specific valence-arousal annotations. Moreover, it operates at the dialogue level and does not account for whether consecutive utterances belong to the same speaker; a speaker-aware extension is left for future work.
Finally, evaluation is restricted to two established benchmarks,
and the updated feature extractors prevent a fully controlled comparison with
the original SDT implementation because its fine-tuned encoder weights were not publicly available.

%% file: sections/5_conclusion.tex
\section{Conclusion}

This work presents a multimodal Transformer-based ERC framework with
geometry-enhanced visual representations, class-wise adaptive modality
fusion, and a valence-arousal prior for emotion transitions. Experiments on
MELD and IEMOCAP show that structured facial descriptors can complement
appearance features, with particularly clear improvements on IEMOCAP.
Class-wise fusion also improves over the original softmax gate and
class-agnostic variants, indicating that modality relevance
depends on the emotion category being predicted. The valence-arousal prior
provides modest but targeted gains on emotionally shifted utterances while
largely preserving performance on stable dialogue turns.

Future research directions include learning dataset- and speaker-specific
affective geometries, improving robustness to degraded modality inputs, and
evaluating the method on more spontaneous conversational data. Another
promising direction is to analyze the learned class-wise weights
at the utterance level to clarify which modality drives each emotion prediction.